\documentclass[twocolumn]{article}
\usepackage[utf8]{inputenc}
\usepackage[inline]{enumitem}
\usepackage{booktabs}
\usepackage{multirow}
\usepackage{url}
\usepackage{caption}
\usepackage{subcaption}
\usepackage{graphicx} 
\usepackage{multicol}
\usepackage{amsmath}
\usepackage{authblk}
\usepackage[round]{natbib}

\begin{document}

\title{Spanish Biomedical and Clinical Language Embeddings}
\author[1]{Asier Gutiérrez-Fandiño\footnote{asier.gutierrez@bsc.es}}
\author[1]{Jordi Armengol-Estapé\footnote{jordi.armengol@bsc.es}}
\author[1]{Casimiro Pio Carrino\footnote{casimiro.carrino@bsc.es}}
\author[1]{Ona De Gibert\footnote{ona.degibert@bsc.es}}
\author[1]{Aitor Gonzalez-Agirre\footnote{aitor.gonzalez@bsc.es}}
\author[1]{Marta Villegas\footnote{Corresponding author: marta.villegas@bsc.es}}
\affil[1]{Barcelona Supercomputing Center}

\maketitle

\section*{Abstract}

We computed both Word and Sub-word Embeddings using FastText. For Sub-word embeddings we selected Byte Pair Encoding (BPE) algorithm to represent the sub-words.

We evaluated the Biomedical Word Embeddings obtaining better results than previous versions showing the implication that with more data, we obtain better representations.

\section{Introduction}
BERT-like \citep{devlin2019bert} and GPT-like \citep{brown2020language} Language Models' effectiveness is corroborated for most of the Natural Language Processing tasks; however, computing more traditional embeddings is still useful as, for tasks with small data and/or scenarios not using large computational resources, they are still competitive. 


These new embeddings use a new Spanish Biomedical Corpus, a Spanish Clinical Corpus, and the use of BPE embeddings. We explain the process of generating the embeddings from two unprecedented Spanish corpora of health. First, we describe the data and the cleaning process, then we explain the embedding methods and, finally, we report the evaluation results.

\section{Corpora}
We have developed two types of embeddings using two different corpora: the Spanish Biomedical Corpora and the Spanish Clinical Corpora. Since the Spanish Biomedical Corpora is of a much larger magnitude in size than the Clinical Corpora, we decided to compute embeddings separately and provide them as distinct resources.

\subsection{Spanish Biomedical Corpora}
We used a big biomedical corpora gathering from a variety of medical resources, namely scientific literature, clinical cases and crawled data.

\begin{table}[]
\centering
\begin{tabular}{@{}lr@{}}
\toprule
Resource name & \multicolumn{1}{l}{Size (tokens)} \\ \midrule
Medical crawler & 746,368,185 \\ 
Books\_general & 97,146,139 \\
Scielo & 60,007,289 \\
BARR2\_background & 24,516,442 \\
Wikipedia\_life\_sciences & 13,890,501 \\
Patents & 13,463,387 \\
EMEA & 5,377,448 \\
REEC & 4,283,453 \\
Mespen\_Medline & 4,166,077 \\
Pubmed & 1,858,966 \\
Books of clinical cases & 1,024,797 \\
Radiology clinical cases & 170,997 \\
Cardiology clinical cases & 147,790 \\
Covid clinical cases & 82,091 \\
\bottomrule
\end{tabular}
\caption{Biomedical Corpora (Bio-Corpus) resources.}
\label{tab:spanish-bio-size}
\end{table}

Table \ref{tab:spanish-bio-size} shows the composition of the largest Spanish Biomedical Corpora ever made. The corpus includes: cardiology clinical cases, radiology clinical cases, clinical cases books, COVID clinical cases, EMEA clinical cases\footnote{EMEA is a corpus of biomedical documents retrieved from the European Medicines Agency (EMEA). The corpus includes documents related to medicinal products and their translations into 22 official languages of the European Union.} \citep{TIEDEMANN12.463}, medical patents, Life Sciences Wikipedia download, barr2\_background \citep{conf/sepln/IntxaurrondoMGL18}, PubMed data, Reec\footnote{Registro español de estudios clinicos: \url{https://reec.aemps.es/reec/public/web.html}}, Medline data, General PDFs, Scielo data and a large medical crawling \citep{krallinger_martin_2021_4561971}.  See \citep{Villegas2018TheMR} for further details on MeSpEn resource, which compiles many of the previously mentioned corpora.

\begin{table*}[ht!]
\centering
\begin{tabular}{@{}lllllll@{}}
 &  &  & \multicolumn{2}{c}{\textbf{Cased}} & \multicolumn{2}{c}{\textbf{Uncased}} \\ \cmidrule(l){4-7} 
\textbf{Version} & \textbf{Method} & \textbf{Corpora} & \textbf{Validation} & \textbf{Test} & \textbf{Validation} & \textbf{Test} \\ \midrule
\multirow{3}{*}{v1.0} & \multirow{3}{*}{Skip-gram} & Wiki & 88.55 & 87.78 & - & - \\
 &  & SciELO & 89.47 & 87.31 & - & - \\
 &  & SciELO+Wiki & 89.42 & 88.17 & - & - \\ \midrule
\multirow{6}{*}{v2.0} & \multirow{3}{*}{CBOW} & Wiki & 86.55 & 85.46 & 86.70 & 86.34 \\
 &  & SciELO & 88.11 & 87.75 & 86.99 & 87.58 \\
 &  & SciELO+Wiki & 88.68 & 86.58 & 86.65 & 85.27 \\
 & \multirow{3}{*}{Skip-gram} & Wiki & 88.62 & 87.16 & 88.31 & 87.43 \\
 &  & SciELO & 89.66 & 88.77 & 89.57 & 89.61 \\
 &  & SciELO+Wiki & 88.76 & 88.64 & 89.82 & 88.28 \\ \midrule
\multirow{2}{*}{v3.0} & CBOW & \multirow{2}{*}{Bio-Corpus} & 88.92 & 88.12 & 88.86 & 88.41 \\
 & Skip-gram &  & 90.91 & \textbf{89.40} & 89.97 & \textbf{89.66} \\ \bottomrule
\end{tabular}
\caption{Bio-Corpus embeddings (v3.0) compared to previous versions of the Spanish Biomedical Word Embeddings.}
\label{tab:biomedical-evaluation}
\end{table*}

\subsection{Spanish Clinical Corpora}
The clinical Corpora is conformed by 5 main corpora, the information contained by these corpora are mainly COVID-19 cases and ictus cases. 

\subsection{Cleaning}
The source of the corpora of both biomedical and clinical domains is of multiple typologies: PDF, WARCs, plain text, etcetera. We cleaned each corpus independently applying a cleaning pipeline with customized operations designed to read data in different formats, split into sentences, perform language detection, remove noisy and malformed sentences, deduplicate and eventually output the data with their original document boundaries. Finally, in order to avoid repetitive content, we concatenated all the individual corpora and deduplicated again common documents among them.

\section{Methods}
We provide two type of embeddings: FastText Word Embeddings and BPE Sub-word Embeddings. 

\subsection{FastText Embeddings}
FastText embeddings are explained in \cite{bojanowski2017enriching}. We tokenized the sentences and used the script available in the website\footnote{\url{https://fasttext.cc/docs/en/unsupervised-tutorial.html}}. As embedding size we used 50, 100 and 300 dimensions. For embedding methods, we used CBOW and Skip-gram. For the Biomedical Corpora, we set the minimum threshold for word frequency to 1 but for the Clinical Corpora we increased the threshold to 4 to avoid leaking sensitive data.

\subsection{BPE Embeddings}
BPE embeddings are introduced in \cite{heinzerling2018bpemb}. The vocabulary size parameter controls the sub-word splitting mechanism. We set the vocabulary size to 8,000 in the case of the Clinical domain and 10,000 in the case of the Biomedical domain. For the uncased version, before computing the BPE vocabulary, the corpus is lower cased. After computing the BPE subwords, FastText embeddings are computed using the official script, omitting the word threshold in the clinical corpus.

\section{Evaluation}
We evaluated the biomedical embeddings using the same scenario of a previous work \citep{soares-etal-2019-medical} using the PharmaCoNER dataset \citep{gonzalez-agirre-etal-2019-pharmaconer}. Clinical Word Embeddings and Sub-word embeddings are not evaluated due to the lack of a suitable evaluation scenario. Table \ref{tab:biomedical-evaluation} shows the results.

With this new version (v3.0) of the Biomedical Word Embeddings, we obtain almost 91\% and 90\% in both cased and uncased validation sets and the Skip-gram embedding method. In the test set we obtain almost 90\% in both cased and uncased test sets using the Skip-gram embedding method. We improve the results of previous versions of the embeddings in both validation and test sets.

\section{Conclusions \& Future Work}
In this work we provide new materials for the Natural Language Processing community regarding medical domain in Spanish. With these resources we aim to fill the lack of resources in medical AI due to the sensitivity of medical data.

We explained how our corpora is conformed, the embedding methods we used and the evaluation we followed.

We have shown that with larger corpus, our embeddings capture more information to accomplish the evaluation providing better results. Our embeddings show a steady improvement as more corpora have been available.

\section*{Materials}
All the Embeddings have been uploaded to Zenodo:
\begin{itemize}
    \item Biomedical Word Embeddings: \url{https://zenodo.org/record/4543236}
    \item Biomedical Sub-word Embeddings: \url{https://zenodo.org/record/4557459}
    \item Clinical Word Embeddings: \url{https://zenodo.org/record/4552042}
    \item Clinical Sub-word Embeddings: \url{https://zenodo.org/record/4555598}
\end{itemize}

\section*{Acknowledgements}
This work has been partially funded by the State Secretariat for Digitalization and Artificial Intelligence (SEDIA) to carry out specialised technical support activities in supercomputing within the framework of the Plan TL\footnote{\url{https://www.plantl.gob.es/}} signed on 14 December 2018; and the ICTUSnet INTERREG Sudoe programme.

\medskip

\bibliographystyle{plainnat}
\bibliography{references} 

\end{document}